\documentclass[3p,authoryear]{elsarticle}

\usepackage{amssymb}
\usepackage{pifont}

\newcommand{\xmark}{\ding{55}}

\usepackage{algorithm}
\usepackage{algorithmic}

\usepackage[retain-explicit-plus]{siunitx}
\usepackage{lineno,hyperref}
\usepackage{mathtools}
\usepackage{multirow}
\usepackage{multicol}
\usepackage{booktabs}
\usepackage{listings}
\usepackage{makecell}
\usepackage{tablefootnote}
\usepackage{subcaption}
\usepackage{threeparttable}
\usepackage[labelsep=quad]{caption}
\usepackage{setspace}
\usepackage{xcolor}
\usepackage{footmisc}
\usepackage[font=normalsize]{caption}
\usepackage{makecell}

\usepackage{blindtext}
\usepackage{hyperref}
\usepackage{nameref}

\newcounter{mylabelcounter}

\makeatletter
\newcommand{\labelText}[2]{%
\refstepcounter{mylabelcounter}%
\immediate\write\@auxout{%
  \string\newlabel{#2}{{\unexpanded{#1}}{\thepage}{{\unexpanded{#1}}}{mylabelcounter.\number\value{mylabelcounter}}{}}%
}%
}
\makeatother

\usepackage{cleveref}
\journal{Applied Artificial Intelligence}
\biboptions{authoryear}

\makeatletter
\def\ps@pprintTitle{%
 \let\@oddhead\@empty
 \let\@evenhead\@empty
 \let\@oddfoot\@empty
 \let\@evenfoot\@empty
}
\makeatother

\begin{document}

\begin{frontmatter}
\lstset{basicstyle=\normalsize\ttfamily,breaklines=true}

\title{Detecting and explaining postpartum depression in real-time with generative artificial intelligence\footnote{García-Méndez, S., \& de Arriba-Pérez, F. (2025). Detecting and Explaining Postpartum Depression in Real-Time with Generative Artificial Intelligence. Applied Artificial Intelligence, 39(1), 2515063. Version of record at: https://www.tandfonline.com/doi/full/10.1080/08839514.2025.2515063.}}

\author[mymainaddress]{Silvia García-Méndez\corref{mycorrespondingauthor}}
\ead{sgarcia@gti.uvigo.es}
\author[mymainaddress]{Francisco de Arriba-Pérez}
\ead{farriba@gti.uvigo.es}

\address[mymainaddress]{Information Technologies Group, atlanTTic, University of Vigo, Vigo, Spain}

\cortext[mycorrespondingauthor]{Corresponding author: sgarcia@gti.uvigo.es}

\begin{abstract}
Among the many challenges mothers undergo after childbirth, postpartum depression (\textsc{ppd}) is a severe condition that significantly impacts their mental and physical well-being. Consequently, the rapid detection of \textsc{ppd} and their associated risk factors is critical for in-time assessment and intervention through specialized prevention procedures. Accordingly, this work addresses the need to help practitioners make decisions with the latest technological advancements to enable real-time screening and treatment recommendations. Mainly, our work contributes to an intelligent \textsc{ppd} screening system that combines Natural Language Processing, Machine Learning (\textsc{ml}), and Large Language Models (\textsc{llm}s) towards an affordable, real-time, and non-invasive free speech analysis. Moreover, it addresses the black box problem since the predictions are described to the end users thanks to the combination of \textsc{llm}s with interpretable \textsc{ml} models (\textit{i.e.}, tree-based algorithms) using feature importance and natural language. The results obtained are \SI{90}{\percent} on \textsc{ppd} detection for all evaluation metrics, outperforming the competing solutions in the literature. Ultimately, our solution contributes to the rapid detection of \textsc{ppd} and their associated risk factors, critical for in-time and proper assessment and intervention.
\end{abstract}

\begin{keyword}
eXplainable and generative Artificial Intelligence, healthcare case study, human-computer interaction, Large Language Models, Natural Language Processing, postpartum depression, stream-based Machine Learning
\end{keyword}

\end{frontmatter}


\section{Introduction}

Depression is a global public health concern that affects more than 150 million people, being more prevalent in women \citep{Labaka2018,Moreira2019}. Notably, the mental and physical well-being of women is significantly affected by pregnancy and ultimately with childbirth when the levels of reproductive hormones decline rapidly (\textit{e.g.}, they experience anxiety, appetite disturbance, insomnia, irritability, lack of concentration, mood disorder, stress) \citep{Yang2022,su2023}. Among the many challenges mothers undergo after childbirth, postpartum depression (\textsc{ppd}) is a severe condition that usually requires medical intervention \citep{Falana2019}.

Mainly, \textsc{ppd} is a common non-psychotic mental disorder during the first year after childbirth that can lead to severe complications in the women's health \citep{Abadiga2019}. Current data indicates that between \SI{10}{\percent} to \SI{15}{\percent} of mothers worldwide are affected with \textsc{ppd} yearly \citep{Fatima2019,Liu2023}. Moreover, only \SI{20}{\percent} of the target population is diagnosed or even treated promptly \citep{Mazumder2021}. Sadly, their cognitive and emotional state directly affects the newborns regarding mother-child attachment and their proper development \citep{Slomian2019,Rogers2020,Andersson2021}.

Consequently, the rapid detection of \textsc{ppd} and their associated risk factors is critical for in-time assessment and intervention through specialized procedures, even more so for vulnerable women \citep{Wang2019}. Unfortunately, the early detection of \textsc{ppd} is compromised by the current solutions \citep{Nurbaeti2021}. The latter can be attained with cost-efficient and effective systems, such as intelligent assistants or chatbots \citep{dergaa2024chatgpt}. However, their application in mental health diagnosis is relatively new \citep{Cameron2019,Duvvuri2022}.

Notably, despite the current technological advances, \textsc{ppd} screening continues to be performed mainly using the traditional survey-based methods (\textit{e.g.}, Edinburgh Postnatal Depression Scale - \textsc{epds}, Patient Health Questionnaire-2/9 - \textsc{phq}-2/9) which rely on self-report (\textit{i.e.}, subjective data) gathered from the prenatal period. Unfortunately, the results of the latter procedure may not reflect the highly complex pathogenesis of \textsc{ppd}, which derives from environmental, genetic, hormonal, and psychological factors \citep{Moreira2019,Stewart2019} or may be biased by lack of awareness, and stigma among other relevant factors \citep{Gabrieli2020}.

In contrast, Machine Learning (\textsc{ml}) has the potential to help healthcare practitioners detect at-risk patients by assisting them during decision-making through precise estimates inferred using recent data from different sources (\textit{e.g.}, informal notes and electronic health records - \textsc{ehr}s) \citep{Shatte2019,borsos2024predicting}. Particularly, \textsc{ml} algorithms can model a vast amount of multidimensional and non-linear data thanks to their enhanced statistical capabilities compared to the traditional linear analysis approach \citep{Liu2023}. In fact, \textsc{ml} is used more frequently in personalized health care, particularly in the mental health field \citep{Dwyer2018,Graham2019,Tai2019}. The traditional \textsc{ml} techniques (\textit{i.e.}, supervised, semi-supervised, unsupervised) can be applied in offline (batch processing) or online (stream processing) \citep{Monteiro2021}. The former batch processing creates fixed models from the training and testing partitions of the experimental data. In contrast, the latter stream processing builds incremental models in real-time, which results in an up-to-date knowledge base \citep{Kolajo:2019}. Thus, we are especially interested in online detection of \textsc{ppd}.

Furthermore, \textsc{ml} can benefit from Natural Language Processing (\textsc{nlp}) techniques to take the most from the human-language data, \textit{i.e.}, extract relevant and high-level reasoning features and translate them in the most appropriate way to computer-interpretable knowledge \citep{sim2023natural}. Unfortunately, the latest technological advancements, in terms of Large Language Models (\textsc{llm}s), are not fully exploited in detecting and treating \textsc{ppd} yet \citep{Liu2023}. In this line, \textsc{llm}s leverage the massive potential of deep learning techniques that can be highly useful in medical practice \citep{jethani2023evaluating,andargoli2024intelligent}. 

Finally, it should be noted that the vast majority of the current approaches and techniques for \textsc{ppd} detection are affected by the nowadays popular black box issue. Explaining how the intelligent system has come to a particular prediction may significantly reduce the black box issue for developers and end users. The latter can be specifically positive in healthcare to generate corrective procedures \citep{Agbavor2022}. To fight it, the research community should favor interpretable models when possible (\textit{e.g.}, tree-based models like Random Forest - \textsc{rf}) or create \textit{ad hoc} modules to circumvent it by exploiting eXplainable Artificial Intelligence (\textsc{xai}) \citep{BarredoArrieta2020}. This is highly appropriate to the health care field, where understanding the decision-making process of the diagnosis may be crucial and benefit personalized treatments \citep{Agbavor2022,gondocs2024ai}.

Thus, an opportunity exists to contribute to the current state of the art on \textsc{ppd} detection using the latest technology. Consequently, our work contributes to an intelligent \textsc{ppd} screening system from spontaneous speech that combines \textsc{nlp}, \textsc{ml}, and \textsc{llm}s towards an affordable, real-time, and non-invasive procedure. Moreover, it addresses the black box problem since the predictions are explained with feature importance and natural language.

The rest of this paper is organized as follows. Section \ref{sec:related_work} overviews the relevant competing \textsc{ppd} solutions in the state of the art, paying particular attention to applying advanced techniques. Section \ref{sec:system} introduces the proposed solution, while Section \ref{sec:results} describes the experimental dataset, implementations, and setup, along with the empirical evaluation results. Finally, Section \ref{sec:conclusion} concludes and highlights the achievements and future work.

\section{Related Work}
\label{sec:related_work}

\textsc{ai}-based healthcare systems enable affordable, non-invasive, or friendly and rapid screening \citep{Yu2018,aminizadeh2024opportunities}. Our research mainly focuses on using \textsc{llm}s in the health care field. Note that \textsc{llm}s exhibit great Natural Language Generation (\textsc{nlg}) and Natural Language Understanding (\textsc{nlu}) functionalities \citep{Agbavor2022} that can be highly appreciated in medical practice \citep{jethani2023evaluating}. 

In this line, \textsc{gpt-3.5}, \textsc{gpt-4}, and \textsc{gpt-4}o are the models used in the nowadays popular Chat\textsc{gpt} assistant with the aforementioned \textsc{nlg} and \textsc{nlu} capabilities improved compared to \textsc{gpt-3} and other earlier models. Specific applications of the latter intelligent assistant exist in the literature to generate clinical notes \citep{Cascella2023} and provide medical advice \citep{Ayers2023}. However, \textsc{llm}s concentrate nowadays on semantics learning, \textit{i.e.}, general \textsc{nlp} capabilities. Thus, they can be further enhanced when combined with \textsc{ml} to provide them with specific knowledge, in our case, medical data, while overcoming the lack of real-time data in their knowledge base \citep{Liao2023}. 

Unfortunately, \textsc{llm}s are not fully exploited in detecting and treating \textsc{ppd} yet \citep{Liu2023}. A preliminary but representative exception is the work by \citet{sezgin2023clinical}. The authors evaluated \textsc{gpt-4} (employing Chat\textsc{gpt}) and \textsc{l}a\textsc{mda} (using Bard) when answering common questions related to \textsc{ppd}. Board-certified physicians assessed their clinical accuracy, and Chat\textsc{gpt} attained the best results.

Notwithstanding the lack of studies exploiting the aforementioned \textsc{llm} models, there exists solid research in \textsc{ppd} detection using \textsc{ml}, some of which exploit textual input data. \citet{Moreira2019} exploited clinical, demographic, and social data to predict the risk of \textsc{ppd} in women affected with hypertension during their pregnancy. Similarly, \citet{Andersson2021} and \citet{Zhang2021} used clinical and demographic features in \textsc{ml} models for \textsc{ppd} detection. The first work computed feature importance to provide interpretability to some extent. Conversely, \citet{Prabhashwaree2022} identified family- and social-related risk factors for \textsc{ppd} from a survey using \textsc{ml}. In this line, \citet{Raisa2022} compared different \textsc{ml} models towards effective \textsc{ppd} detection for the specific case of the Bangladesh women population. Demographic and social data were gathered manually through an \textit{ad hoc} survey to analyze the correlation between them and \textsc{ppd}. In contrast, \citet{Yang2022} developed a clinical and demographic-based solution for \textsc{ppd} screening for women who underwent a cesarean delivery. They proved the effectiveness of Ketamine intervention to lower the incidence of \textsc{ppd}. More recently, \citet{nakamura2024early} exploited demographic information and subjective ratings to assess \textsc{ppd} with \textsc{ml}.

Regarding works that address \textsc{ppd} analyzing text through \textsc{ml}, \citet{Wang2019,Hochman2021,Zhang2021} used \textsc{ehr}s data, while \citet{Fatima2019} exploited linguistic features extracted from social media posts from Reddit. A similar approach to the latter solution was followed by \citet{Shatte2020}. In this line, \citet{Trifan2020} built a corpus of \textsc{ppd}-related posts from Reddit and proposed a preliminary \textsc{ml}-based solution.

Furthermore, a hybrid system was presented by \citet{Shin2020}. In this studio, the authors relied on lifestyle and maternal demographic features apart from two questions adapted from the \textsc{phq-2} questionnaire about the current emotional state of the women to detect \textsc{ppd}. Moreover, \citet{Mazumder2021} detected \textsc{ppd} with \textsc{ml} models but exploited bagging and boosting techniques using as input data responses to a survey regarding their background and state of mind. 

Apps for \textsc{ppd} screening exist, but without \textsc{nlp} capabilities. It is the case of the work by \citet{Nurbaeti2021}, who developed an Android app to detect early symptoms of \textsc{ppd} through a survey.

Even though outside the \textsc{ppd} research field, but related to our aim of spontaneous speech analysis is the work by \citet{Duvvuri2022} who developed a chatbot for depression detection using data gathered from Discord in an \textsc{ml}-based solution. Similarly, \citet{dergaa2024chatgpt} analyzed the appropriateness of Chat\textsc{gpt} treatment recommendations in different mental health scenarios (\textit{e.g.}, anxiety related to educational training), including \textsc{ppd}. Unfortunately, the authors were limited to assessing the potential of Chat\textsc{gpt} as a healthcare professional's collaborative tool, and no adjustment of its performance through prompt engineering was conducted. Moreover, the recommendations were not justified through \textsc{xai}.

After an extensive revision of the \textsc{ppd} state of the art, we could not detect works that fight the already mentioned black box problem or analyze \textsc{ppd} in streaming despite its suitability due to the vast amount of continuous data available from clinical records. Outside the target field of this work, the research community addressed the responsible use of \textsc{ai} potential through feature importance \citep{FranciscoDeArriba2022}, natural language \citep{Ehsan2019}, and visual explanations \citep{Spinner2019}. 

The few exceptions of the application of \textsc{xai} for \textsc{ppd} are the studies performed by \citet{Amit2021}, \citet{Liu2023}, \citet{qi2025prediction} and \citet{zhang2025interpretable}, all exploited SHapley Additive exPlanations (\textsc{shap}). Firstly, \citet{Amit2021} proposed an \textsc{ml}-based solution trained with \textsc{ehr}s combined with the \textsc{epds}. Moreover, \citet{Liu2023} created an \textsc{ml}-based calculator of \textsc{ppd} risk. However, they only consider women who underwent cesarean delivery, and the features involved are reduced to demographic data. As previously mentioned, the latter may result in input data bias and inhibitions in self-reporting introduced involuntarily by the women involved in the study \citep{Duvvuri2022}. More recently, \citet{qi2025prediction} evaluated the performance of classical \textsc{ml} models for \textsc{ppd} prediction. The authors used biopsychosocial features. Ultimately, \citet{zhang2025interpretable} exploited \textsc{emr}s to train different \textsc{ml} classifiers.

Surprisingly, despite the severity of \textsc{ppd} in our society, the number of current studies on this topic is scarce when it comes to the newest \textsc{ai} techniques (\textit{e.g.}, \textsc{llm}s, \textsc{xai}, online processing) \citep{Andersson2021,Raisa2022}. Accordingly, this work addresses the need to help practitioners in decision-making with the latest technological advancements to enable real-time screening and treatment recommendations. Mainly, our work contributes to an intelligent \textsc{ppd} screening system that combines \textsc{nlp}, \textsc{ml}, and \textsc{llm}s towards an affordable, real-time, and non-invasive analysis of free speech. Moreover, it addresses the black box problem since the predictions are described to the end users thanks to the combination of \textsc{llm}s with interpretable \textsc{ml} models (\textit{i.e.}, tree-based algorithms) using feature importance and natural language. Ultimately, it contributes to the rapid detection of \textsc{ppd} and their associated risk factors, which are critical for in-time and proper assessment and intervention.

\subsection{Contributions}

In light of the comparison with the most closely related works in the literature shown in Table \ref{tab:comparison}, our solution is the first to address the research gap of jointly considering online \textsc{ppd} detection with \textsc{llm}s and providing explainability. Even though state-of-the-art technologies are used, the proposed pipeline is an original contribution to this work. Moreover, our research recognizes the importance of transparency in mental health screening, which is vital in clinical settings to foster trust among end users and promote informed decision-making by physicians.

\begin{table*}[!htbp]
\centering
\caption{\textcolor{black}{Comparison of the most related \textsc{ppd} prediction solutions taking into account the domain of application (\textsc{qa}: question answering), approach followed, data processing (offline, online), input data used (\textsc{fs}: free speech, Demogr.: demographics), and explainability capability (Ex., \textsc{fi}: feature importance, \textsc{nl}: natural language).}}
\label{tab:comparison}
\begin{tabular}{lccccc}
\toprule
\textbf{Authorship} & \textbf{Domain} & \textbf{Approach} & \textbf{Processing} & \textbf{Data} & \textbf{Ex.}\\
\midrule

\citet{Amit2021} & \textsc{ppd} & \textsc{ml} & Offline & \textsc{ehr}s + \textsc{epds} & \textsc{shap}\\

\citet{Andersson2021} & \textsc{ppd} & \textsc{ml} & Offline & Clinical + demographic & \textsc{fi}\\

\citet{Duvvuri2022} & Depression & \textsc{ml} & Offline & \textsc{fs} & \xmark\\

\citet{Liu2023} & \textsc{ppd} & \textsc{ml} & Offline & Demographic & \textsc{shap}\\

\citet{sezgin2023clinical} & \textsc{ppd-qa} & Chat\textsc{gpt} & - & \textsc{fs} & \xmark\\

\citet{qi2025prediction} & \textsc{ppd} & \textsc{ml} & Offline & Biopsychosocial & \textsc{shap}\\

\citet{zhang2025interpretable} & \textsc{ppd} & \textsc{ml} & Offline & \textsc{emr} & \textsc{shap}\\\midrule

\textbf{Proposed} & \textsc{ppd} & Chat\textsc{gpt}+\textsc{ml} & Online & \textsc{fs} & \textsc{fi,nl}\\ \bottomrule
\end{tabular}
\end{table*}

\section{System Architecture}
\label{sec:system}

\begin{figure*}
 \centering
 \includegraphics[scale=0.13]{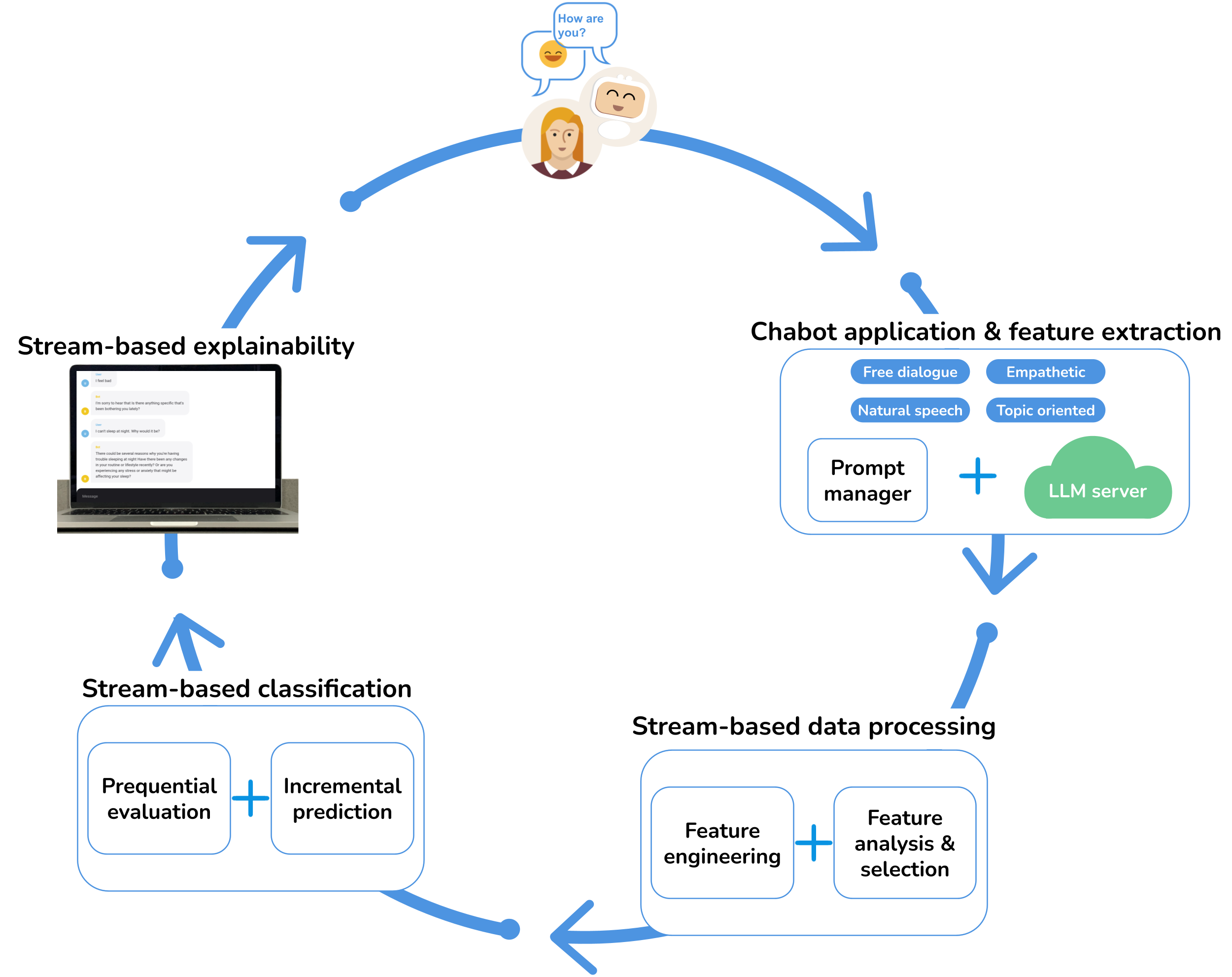}
 \caption{System diagram composed of (\textit{i}) chatbot application \& feature extraction, (\textit{ii}) stream-based data processing involving feature engineering and feature analysis \& selection, (\textit{iii}) stream-based classification, and (\textit{iv}) stream-based explainability modules.}
 \label{fig:scheme}
\end{figure*}

Figure \ref{fig:scheme} shows the scheme of the solution. It comprises a chatbot application (Section \ref{sec:chatbot_feature}) to establish free dialogues with the end users and a feature extraction module that extracts the features through prompt engineering. The latter features are subsequently processed in the stream-based data processing module, which involves feature engineering (Section \ref{sec:feature_engineering}) and feature analysis \& selection (Section \ref{sec:feature_analysis_selection}) and classified by the stream-based classification module (Section \ref{sec:online_classification}). Finally, the most relevant features are exploited to provide explainable descriptions in natural language to the end users (Section \ref{sec:explainability}).

\subsection{Chatbot Application \& Feature Extraction}
\label{sec:chatbot_feature}

The chatbot has been developed as a multi-platform application (\textit{i.e.}, compatible with tablets and smartphones with Android and iOS, and web deployment with macOS and Windows) that enables free and natural dialogues. It can generate empathetic and \textsc{ppd}-oriented utterances through prompt engineering techniques. More in detail, during the conversation with the end user, the chatbot poses questions related to the following topics: (\textit{i}) baby bonding issues, (\textit{ii}) concentration and decision-making problems, (\textit{iii}) feeling sad or tearful and (\textit{iv}) guilty, (\textit{v}) irritability towards the baby or the partner, (\textit{vi}) overreacting or loss of appetite, (\textit{vii}) suicide behavior, and (\textit{viii}) trouble sleeping.

There exist three prompts depending on the analysis objective: (\textit{i}) topic discovery, (\textit{ii}) empathetic text generation by topic, and (\textit{iii}) care treatment proposal. Note that when a topic is already covered in the dialogue, it is removed from the list of topics to avoid repetition. Moreover, the \texttt{temperature} parameter of the \textsc{llm} is exploited towards achieving dialogue naturalness. This parameter enables randomness adjustment of the text generated\footnote{\texttt{Temperature} = 1 means entirely random while 0 stands for deterministic.}. 

\subsection{Feature Engineering}
\label{sec:feature_engineering}

This process is performed by prompt engineering. The latter seeks to interpret the users' responses and translate them into the following options: \texttt{\textsc{na}}\footnote{Not applicable, out of scope.}, \texttt{yes}, \texttt{sometimes}, \texttt{often}, \texttt{no}, \texttt{unwilling to disclose}. Then, this interpretation per topic is associated with a categorical value. Moreover, the chatbot application retrieves the user's age.

Note that the binarization approach for feature engineering is appropriate compared to the transformation into an ordinal scale because the topics covered do not exhibit algebraic relationships among them. The complete process is detailed in Algorithm \ref{alg:get_dummies}. 

\begin{algorithm}
\caption{User response encoding.}
\label{alg:get_dummies}
\begin{algorithmic}[1]
\REQUIRE $dataset$

\FOR{$topic$ in $dataset.columns$}
    \STATE $unique\_values = set(dataset.get\_column(topic))$
    \FOR{$option$ in $unique\_values$}
        \STATE $column\_name = topic + ``\_" + option$
        \STATE $dataset[column\_name].rows = 0$
    \ENDFOR
    \FOR{$index$ in $dataset.get\_column(topic)$}
        \STATE $value = dataset[index][topic]$
        \STATE $column\_name = topic + ``\_" + value$
        \STATE $dataset[index][column\_name] = 1$
    \ENDFOR
    \STATE $dataset.remove\_column(topic)$  
\ENDFOR

\RETURN $dataset$
\end{algorithmic}
\end{algorithm}

\subsection{Feature Analysis \& Selection}
\label{sec:feature_analysis_selection}

The performance of \textsc{ml} models is directly affected by the quality of the experimental data. The latter is especially relevant in the streaming scenario in which samples arrive continuously and must be processed in real time. Moreover, the relevant feature may vary over time in the particular case of feature selection in an online classification problem. Hence, appropriate feature analysis and selection procedures are necessary. 

In this study, the system removes irrelevant features following the variance thresholding methodology. Accordingly, features with a variance value lower than a configurable threshold are discarded for further analysis. The threshold is computed considering the features' variance in a cold start subset of the experimental data and a configurable percentile (see Algorithm \ref{alg:variance}).

\begin{algorithm}
\caption{Variance threshold computation.}
\label{alg:variance}
\begin{algorithmic}[1]
\REQUIRE $dataset$, $percentile$

 \STATE $list\_variances = []$
 
\FOR{$column$ in $dataset.columns$}
    \STATE $variance=dataset[column].var()$
    \STATE $list\_variances.add(variance)$
\ENDFOR

\STATE $variance\_threshold=get\_percentile(percentile,list\_variances)$

\RETURN $variance\_threshold$
\end{algorithmic}
\end{algorithm}

\subsection{Stream-based Classification}
\label{sec:online_classification}

The target variable of our stream-based classification problem is the \textsc{ppd} absence or presence. For that purpose, the following \textsc{ml} models are used:

\begin{itemize}
 
 \item \textbf{Gaussian Naive Bayes} (\textsc{gnb}) \citep{Tieppo2021}. This model is based on the traditional Naive Bayes algorithm, which analyzes Gaussian probability distributions but is configured for stream-based classification. It is exploited as a baseline reference.

 \item \textbf{Logistic Regression} (\textsc{lr}) \citep{Montiel2021}. This model uses a sigmoid function to evaluate the likelihood that a sample belongs to a specific class. As one of the simplest linear classification techniques, it offers fast execution times, although its accuracy may be limited in complex problems.

 \item \textbf{Approximate Large Margin Algorithm} (\textsc{alma}) \citep{Silva2021}. It is similar to the batch version of the Support Vector Machine. It calculates the probability of a category by approximating the maximal margin between a hyperplane and a norm (with a value of $p \geq 2$) for a set of linearly separable data.
 
 \item \textbf{Hoeffding Adaptive Tree Classifier} (\textsc{hatc}) \citep{Mrabet2019}. It is also intended for stream-based classification and exploits a branch performance evaluation mechanism in a single tree structure.
 
 \item \textbf{Adaptive Random Forest Classifier} (\textsc{arfc}) \citep{Fatlawi2020}. This model represents an improvement of the latter \textsc{hatc} algorithm thanks to the ensembled strategy and prediction obtained through majority voting.
 
\end{itemize}

Accuracy, Area Under the Curve (\textsc{auc}), precision, recall, and \textsc{f}-measure are employed for evaluation purposes. While precision assesses the accuracy of optimistic predictions (proportion of true positives among the total positive predictions), the recall metric elaborates on the model's sensitivity to identify all positive instances (proportion of true positives detected over the total number of confirmed positive cases) \citep{Rainio2024}. Thus, a trade-off exists between these two metrics, summarized in the \textsc{f}-measure. A more conservative model is one in which precision is favored over recall \citep{Ho2020}. That is, to ensure the predictions are correct even though not all positive cases are detected. The latter recovery rate represented by the recall can be enlarged, but this may increase the presence of false positives, thus reducing precision. In clinical applications, false positives are those in which a disease is incorrectly identified. The latter can lead to unnecessary treatment that may affect the patient's well-being (\textit{e.g.}, invasive procedures) and subsequent costs. Conversely, false negatives are those cases in which the existing diseases are undetected. The consequences should also be considered (\textit{e.g.}, lack of early and timely treatment). Previous works in the literature maximize precision in uncommon diseases, while recall is favored in serious illnesses \citep{lindsay2024harnessing}.

\subsection{Stream-based Explainability}
\label{sec:explainability}

The explainability data comprises information about the most relevant features extracted to provide valuable insights to the end users in terms of the transparency of the predictions. To extract this information, we apply a technique known as \textit{counterfactual explanation}. It evaluates small perturbations in the features that alter the predicted category. Internally, these permutations produce variations in the decision path followed by the \textsc{ml} model (\textit{i.e.}, \textit{greater than} and \textit{less than} bifurcations in the tree path), enabling changes in the prediction probability. Ultimately, the system considers the features that cause this change to be relevant. The Algorithm \ref{alg:counterfactual} describes the counterfactual explanation procedure. It iterates a finite number of times, searching for the configuration that minimizes the number of affected features.

\begin{algorithm}
\caption{Counterfactual explanation technique.}
\label{alg:counterfactual}
\begin{algorithmic}[1]
\REQUIRE $model$, $x$, $predicted\_label$, $n_{iterations}$ 

\STATE $x\_new=[]$
\STATE $relevant\_features$
\FOR{$i = 1$ to $n_{iterations}$}

    \FOR{each $feature$ in $x$}
        \IF{predicted\_label = 0}
            \STATE $new\_value = random\_choice\_No()$
        \ELSIF{predicted\_label = 1}
            \STATE $new\_value = random\_choice\_Yes()$
        \ENDIF
        \STATE $x\_new[feature]=new\_value$
    \ENDFOR

    \STATE $pred\_new = model.predict\_one(x\_new)$
    \STATE $pred\_proba\_new = model.predict\_proba\_one(x\_new)$

    \IF{$pred\_new != predicted\_label$ and $pred\_proba\_new > 0.5$}
        \STATE $features\_modified=instance\_distance(x, x\_new)$
        \IF{$len(features\_modified) < len(relevant\_features)$}
            \STATE $relevant\_features =  features\_modified$
            \STATE $pred\_proba\_final =  pred\_proba\_new$
            \STATE $x\_final= x\_new$
        \ENDIF
    \ENDIF
\ENDFOR

\RETURN $relevant\_features, pred\_proba\_final, x\_final$
\end{algorithmic}
\end{algorithm}

\section{Evaluation and Discussion}
\label{sec:results}

All the experiments were executed on a computer with the following specifications:
\begin{itemize}
 \item \textbf{Operating System}: Ubuntu 18.04.2 LTS 64 bits
 \item \textbf{Processor}: Intel\@Core i9-10900K \SI{2.80}{\giga\hertz}
 \item \textbf{RAM}: \SI{96}{\giga\byte} DDR4 
 \item \textbf{Disk}: \SI{480}{\giga\byte} NVME + \SI{500}{\giga\byte} SSD
\end{itemize}

\subsection{Experimental Dataset}
\label{sec:experimental_data_Set}

Two datasets\footnote{This study does not involve human participants. Consequently, no ethical approval is needed.} were used: (\textit{i}) a synthetic dataset\footnote{Available at \url{https://doi.org/10.5281/zenodo.14049633}, May 2025.} composed of 200 users' utterances related to the topics described in Section \ref{sec:chatbot_feature} and associated to the users' responses options excluding the \textsc{na} category\footnote{5 utterances per topic and option.}, and (\textit{ii}) a publicly available dataset\footnote{Available at \url{https://www.kaggle.com/datasets/parvezalmuqtadir2348/postpartum-depression}, May 2025.} which comprises \num{1491} users' responses to a medical survey (excluding the \textsc{na} category)\footnote{Note that the users' responses follow the scheme explained before: \texttt{\textsc{na}}, \texttt{yes}, \texttt{sometimes}, \texttt{often}, \texttt{no}, \texttt{unwilling to disclose}.} covering the topics under analysis. The first dataset evaluates the system's \textsc{nlu} capabilities when interpreting the users' responses. In contrast, the second is intended for stream-based \textsc{ppd} classification. 

\begin{figure*}[!htbp]
 \centering
 \includegraphics[scale=0.18]{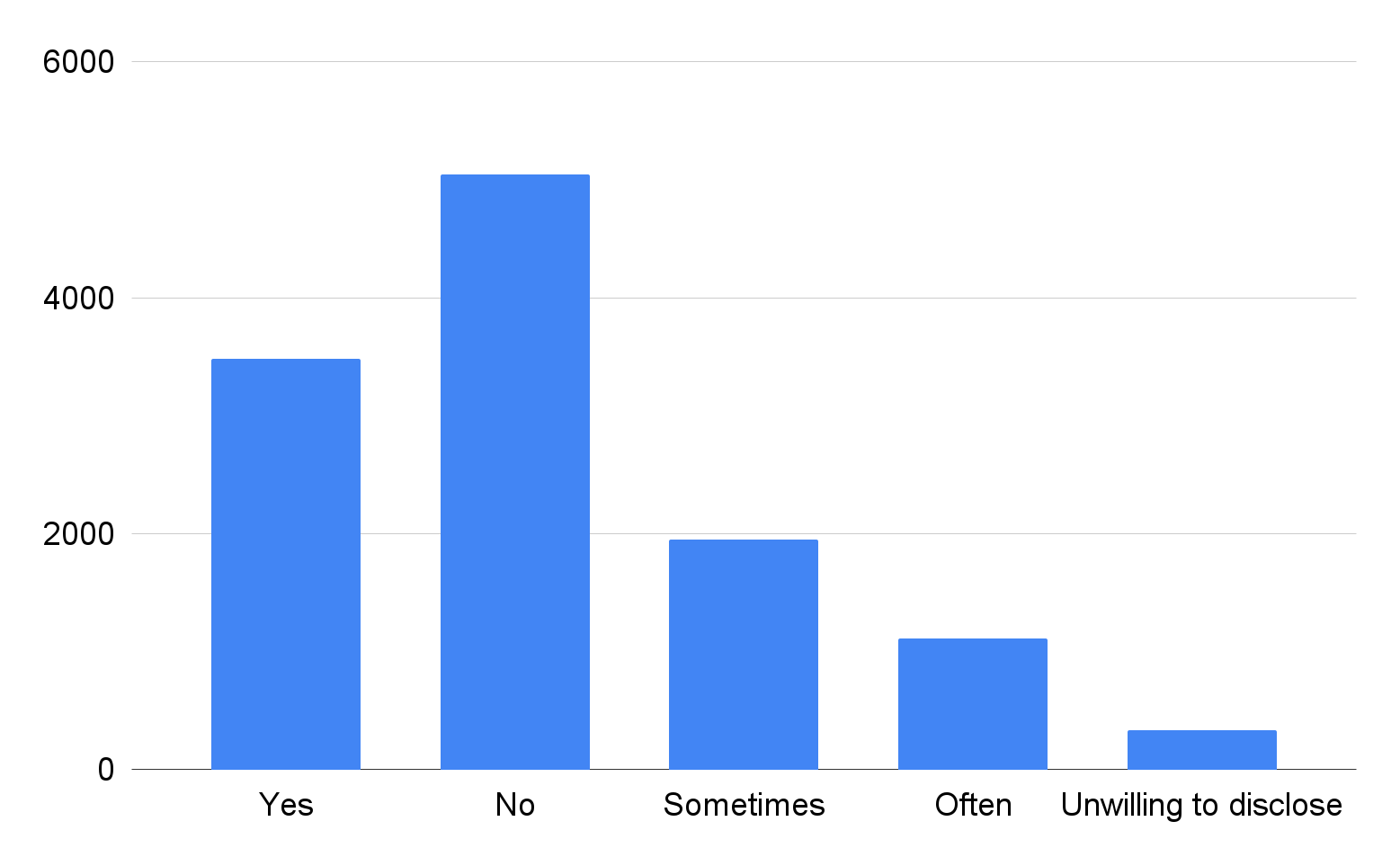}
 \caption{Users' responses distribution.}
 \label{fig:distribution_responses}
\end{figure*}

\begin{figure*}[!htbp]
 \centering
 \includegraphics[scale=0.18]{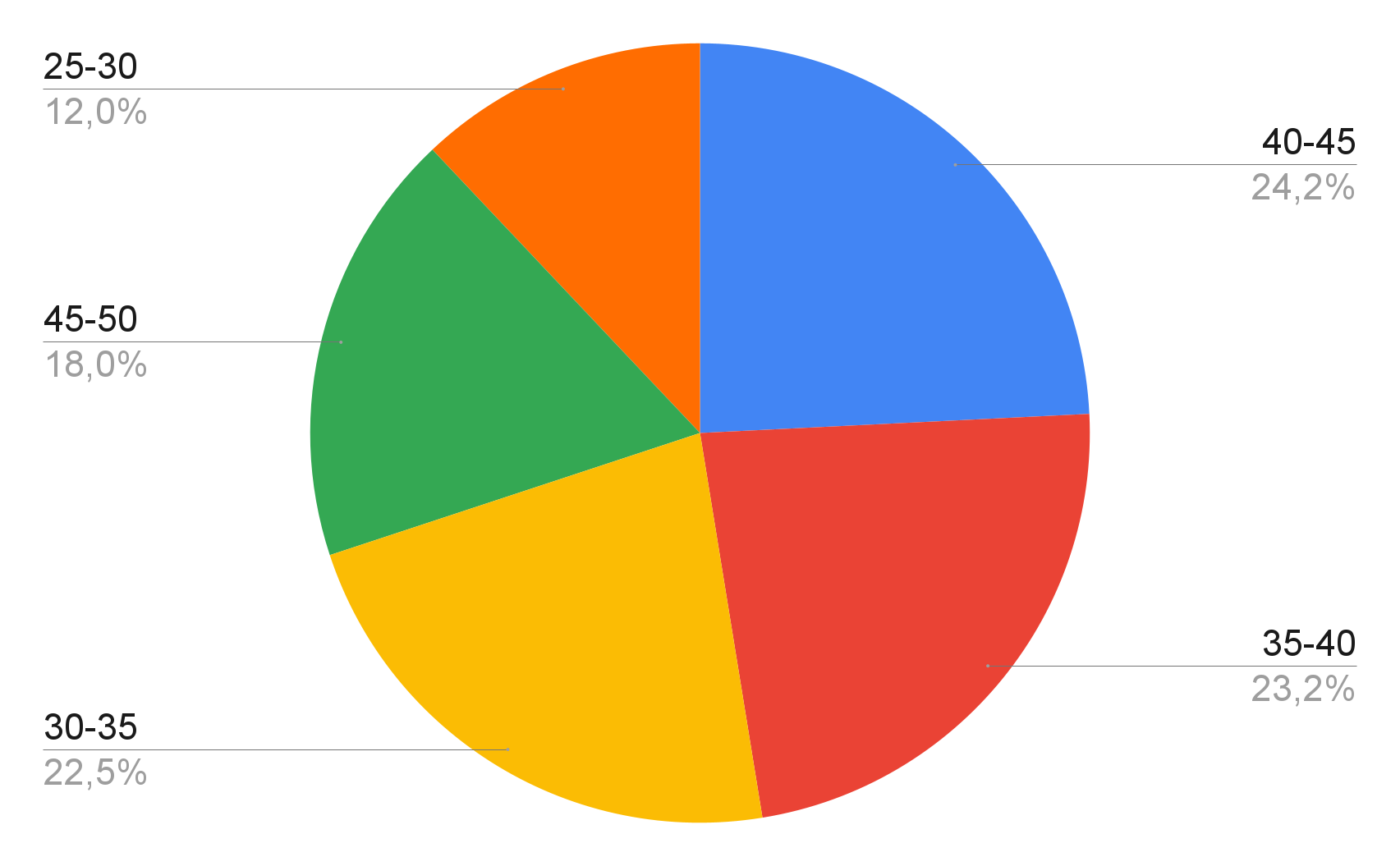}
 \caption{Age distribution in the second dataset.}
 \label{fig:distribution_age_dataset2}
\end{figure*}

The eight symptoms under analysis are equally distributed in the first dataset, 25 utterances each. More in detail, Figure \ref{fig:distribution_responses} details the distribution of users' responses in the second dataset. Moreover, regarding the demographic characteristics of the medical records in the second dataset, the age of the women involved in the study is distributed, as shown in Figure \ref{fig:distribution_age_dataset2}.

Finally, Table \ref{tab:dataset_distribution} shows the distribution of the target variable in the second dataset.

\begin{table}[!htbp]
\centering
\caption{\label{tab:dataset_distribution}Distribution of classes in the second experimental dataset.}
\begin{tabular}{lS[table-format=6.0]}
\toprule \textbf{Category} & \multicolumn{1}{c}{\textbf{Number of samples}}\\ \midrule
Absence of \textsc{ppd} & 523 \\
Presence of \textsc{ppd} & 968 \\
\midrule
Total & 1491 \\ \bottomrule
\end{tabular}
\end{table}

The synthetic dataset was generated to simulate speech patterns and emotional expressions typically observed in postpartum contexts, with special attention to lexical, prosodic, and affective features relevant to depressive symptoms. While it aims to approximate real-world postpartum speech, synthetic data may not fully capture patient data's complexity, nuance, and variability. Therefore, we combined two datasets, one synthetically generated and one publicly available. Regarding the limitations in representativeness or generalizability of the datasets, while it is useful for controlled experiments, the synthetic one may not fully capture the linguistic, emotional, or cultural nuances of real postpartum speech. In contrast, the public dataset may be limited in age, ethnicity, language variety, and socioeconomic diversity, affecting its applicability to broader populations.

\subsection{Chatbot Application \& Feature Extraction}
\label{sec:chatbot_feature_results}

The multi-platform conversational assistant was designed using the Flutter programming framework\footnote{Available at \url{https://flutter.dev}, May 2025.}. The chatbot application was codified in Python (version 3.8)\footnote{Available at \url{https://www.python.org}, May 2025.} in a server implemented in Flask (version 2.2.2)\footnote{Available at \url{https://flask.palletsprojects.com/en}, May 2025.} including a traffic balancer with Gunicorn (version 20.1.0)\footnote{Available at \url{https://gunicorn.org}, May 2025.}, while the Flyer Chat library\footnote{Available at \url{https://flyer.chat}, May 2025.} was used to provide the system with instant messaging capabilities. The chatbot application in the Flask server sends the prompts and the utterances to the \textsc{llm} model used, Chat\textsc{gpt} (version 3.5)\footnote{Available at \url{https://platform.openai.com/docs/models}, May 2025.}, through the Open\textsc{ai} \textsc{rest api}\footnote{Available at \url{https://platform.openai.com}, May 2025.}.

Note that a history of the last 10 interactions is maintained to achieve a fully connected and coherent dialogue. Once this number is exceeded, the oldest interactions are deleted, following the analogy of a \textsc{fifo} stack. Table \ref{tab:prompts} displays the prompts designed and the temperature configuration.

\begin{table*}[!htbp]
\centering
\caption{\label{tab:prompts}Prompt engineering.}
\begin{tabular}{ccc}
\toprule 
\textbf{Prompt ID} & \textbf{Prompt content} & \textbf{Temperature}\\ \midrule
1 & \begin{tabular}[c]{@{}p{7cm}@{}} Analyze the user's responses in the dialogue and return per topic one of the following options: \texttt{\textsc{na}}, \texttt{yes}, \texttt{sometimes}, \texttt{often}, \texttt{no}, \texttt{unwilling to disclose}.
Topic 1: baby bonding issues, topic 2: concentration and decision-making problems, topic 3: feeling sad or tearful, topic 4: feeling guilty, topic 5: irritability towards the baby or the partner, topic 6: overreacting or loss of appetite, topic 7: suicide behavior, topic 8: trouble sleeping.
\end{tabular} & 0\\
\midrule

2 & \begin{tabular}[c]{@{}p{7cm}@{}} The following is a conversation with an \textsc{ai} assistant. The assistant is dynamic, never repeats the same thing twice, and is creative, intelligent, and kind. The assistant tries to establish a conversation, and it always asks a question when he finishes speaking about one of these topics: topic 1: baby bonding issues, topic 2: concentration and decision-making problems, topic 3: feeling sad or tearful, topic 4: feeling guilty, topic 5: irritability towards the baby or the partner, topic 6: overreacting or loss of appetite, topic 7: suicide behavior, topic 8: trouble sleeping. Return an utterance per topic.
\end{tabular} & 1\\
\midrule

3 & \begin{tabular}[c]{@{}p{7cm}@{}} Propose three care treatments for a user with \textsc{ppd} taking into consideration its responses into the following dialogue.
\end{tabular} & 1\\
 \bottomrule
\end{tabular}
\end{table*}

\begin{figure*}
 \centering
 \includegraphics[scale=0.22]{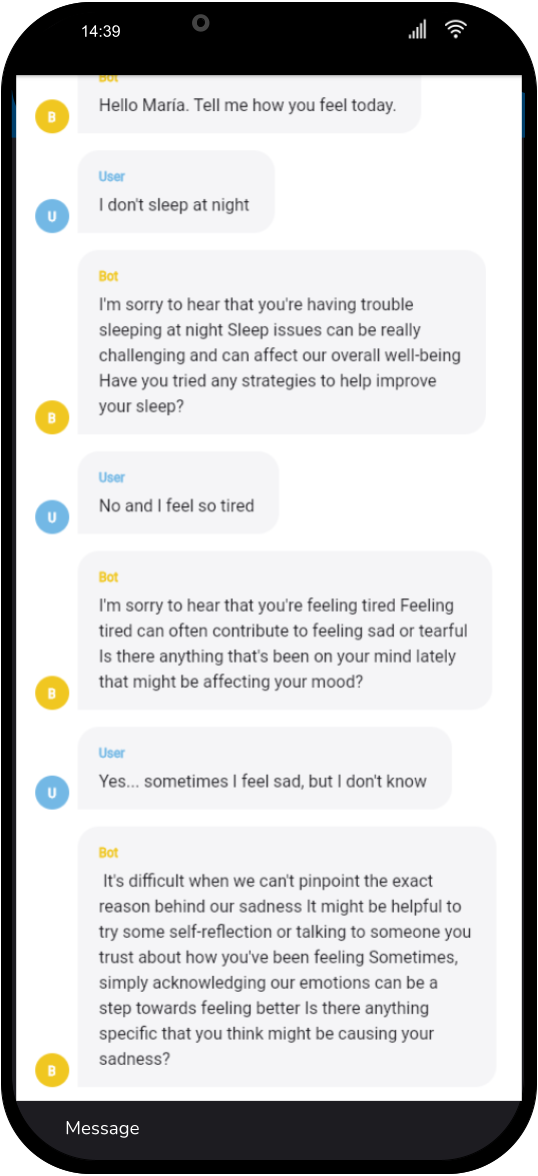}
 \caption{Real conversation example.}
 \label{fig:real_conversation}
\end{figure*}

Moreover, Figure \ref{fig:real_conversation} shows a real user-chatbot dialogue in which the system detected that the user sometimes feels sad, tired, and has trouble sleeping.

\begin{figure*}[!htbp]
 \centering
 \includegraphics[scale=0.25]{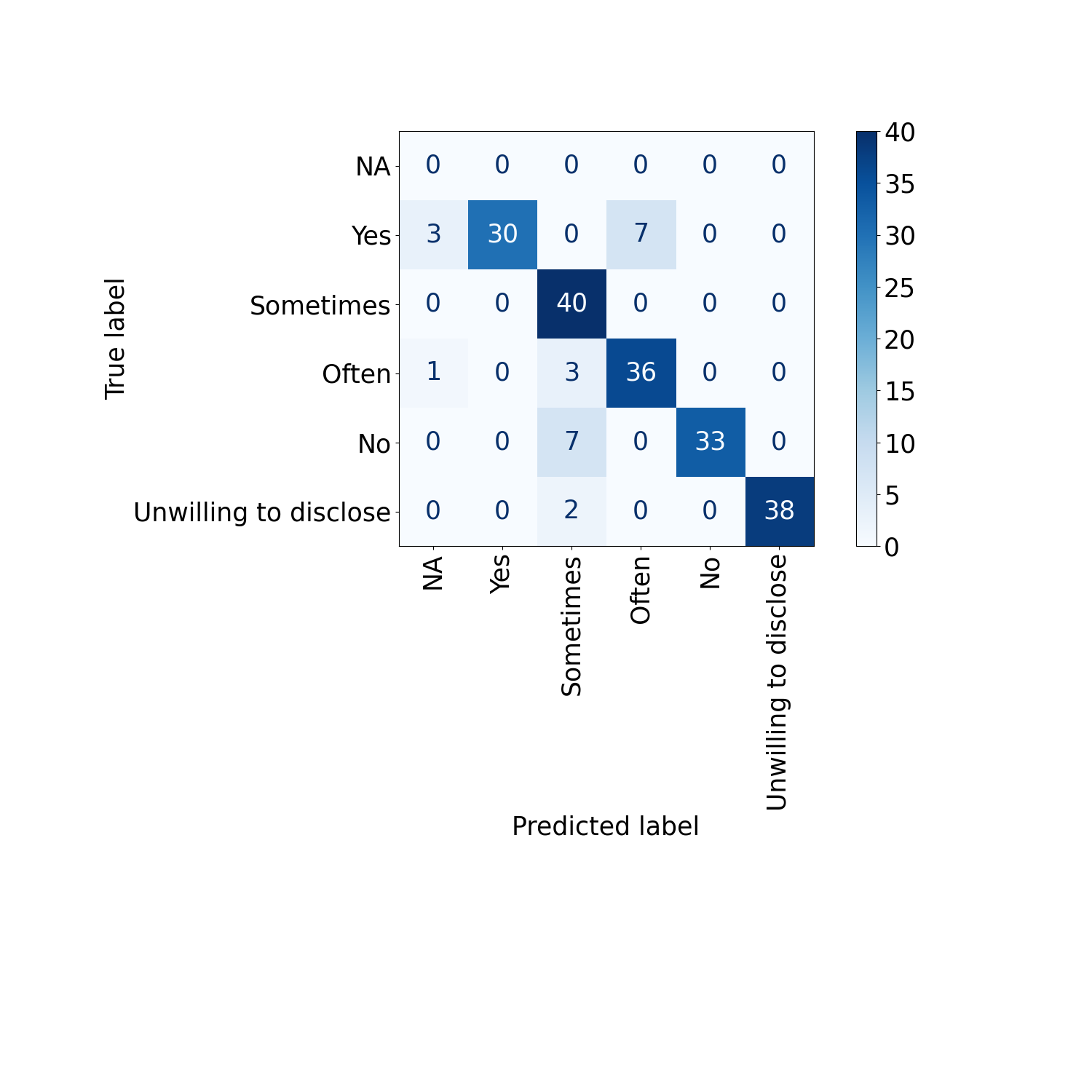}
 \caption{Confusion matrix of the users' responses interpretation evaluation.}
 \label{fig:confusion_matrix}
\end{figure*}

The evaluation of the system's performance when interpreting users' responses exploiting the first dataset achieved an \SI{88.50}{\percent} accuracy. The confusion matrix of this evaluation is shown in Figure \ref{fig:confusion_matrix}. Note that the errors are limited, and only in 4 cases the system failed to comprehend the user's utterance (\textit{i.e.}, relating their response to \textsc{na}). Table \ref{tab:classification_gpt} details the evaluation results with an overall performance around \SI{90}{\percent}. 

\begin{table*}[!htbp]
\centering
\caption{\label{tab:classification_gpt}Users’ responses interpretation results.}
\begin{tabular}{lccccc}
\toprule
 \bf Class & \bf Accuracy &\bf Precision & \bf Recall & \bf F-measure\\
\midrule
Overall & 88.50 & 92.13 & 88.50 & 89.45\\
Yes & - & \bf100.00 & 75.00 & 85.71\\
Sometimes & - & 76.92 & \bf100.00 & 86.96\\
Often & - & 83.72 & 90.00& 86.75\\
No & - & \bf100.00 & 82.50 & 90.41\\
Unwilling to disclose & - & \bf100.00 & 95.00 & \bf97.44\\

\bottomrule
\end{tabular}
\end{table*}

\subsection{Feature Engineering}
\label{sec:feature_engineering_results}

\begin{figure*}[!htbp]
 \centering
 \includegraphics[scale=0.25]{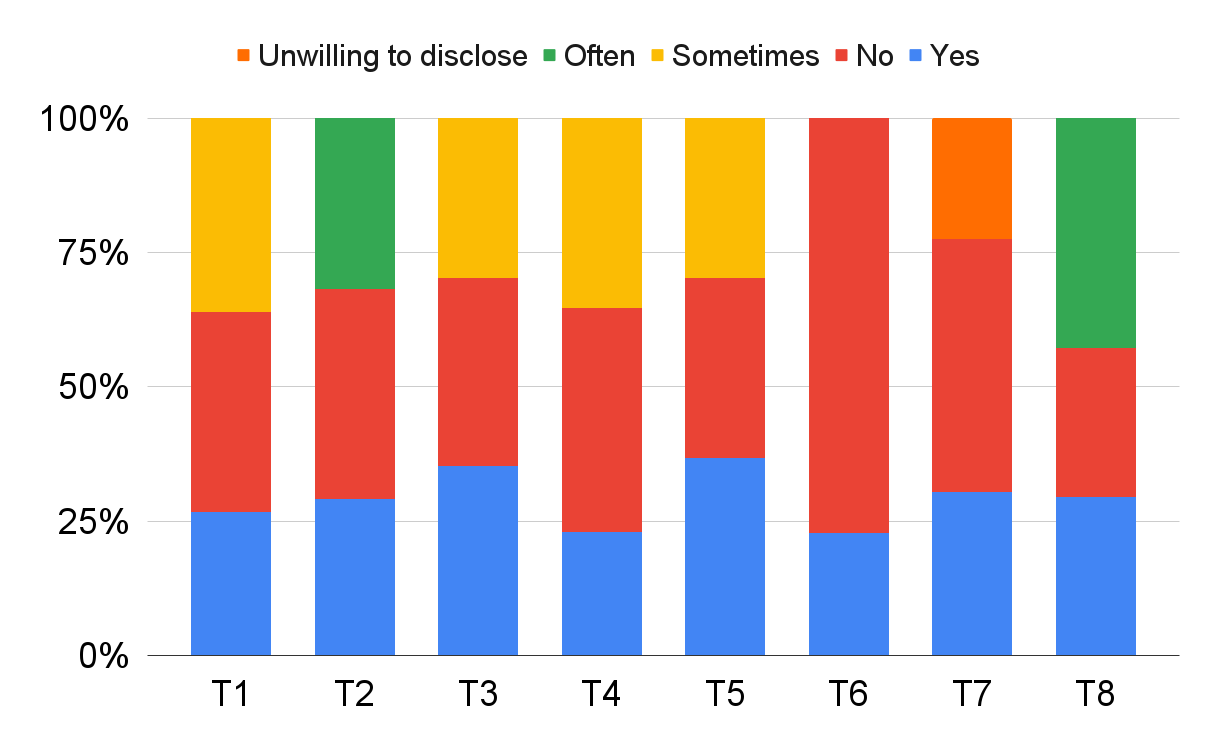}
 \caption{Topic and response distribution.}
 \label{fig:hist_feature_engineering_results}
\end{figure*}

The binarization process uses the \texttt{get\_dummies} function from the Python \texttt{pandas} library\footnote{Available at \url{https://pandas.pydata.org}, May 2025.}. In total, 53 Boolean features are generated, 48 Boolean features corresponding to the 8 topics, 6 options of interpretation of the users' responses, and 5 additional age features\footnote{5-year intervals between 25 and 50 years old.}. In Figure \ref{fig:hist_feature_engineering_results}, we can observe the distribution of the users' responses by the topic under analysis.

\subsection{Feature Analysis \& Selection}
\label{sec:feature_analysis_selection_results}

Note that the variance threshold is established as the 5th percentile value \citep{CastilloT2021} of the variance of the features contained in the \SI{10}{\percent} of the experimental dataset \citep{Cao2022}. The latter represents the cold start setting of the feature selection method. As a result, the threshold was set to \num{0.079} to discard features with low variance fluctuations.

\texttt{VarianceThreshold}\footnote{Available at \url{https://riverml.xyz/0.11.1/api/feature-selection/VarianceThreshold}, May 2025.} method available in the \texttt{River} library\footnote{Available at \url{https://riverml.xyz/0.11.1}, May 2025.} is used to compute the feature variance. 

\begin{figure*}[!htbp]
 \centering
 \includegraphics[scale=0.25]{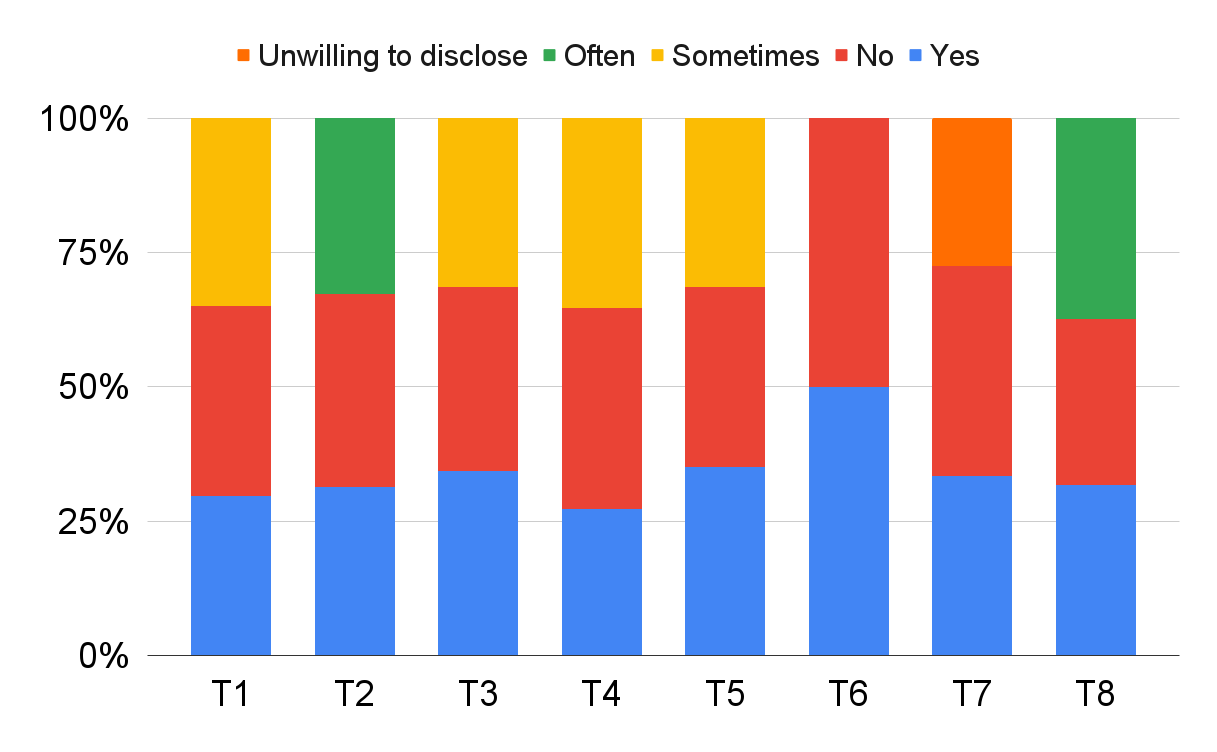}
 \caption{Variance per topic and user response.}
 \label{fig:hist_variance_results}
\end{figure*}

In Figure \ref{fig:hist_variance_results}, we present the variance value of the features in the last processed sample by the online system. As can be observed, all features meet the eligibility criteria and are considered relevant.

\subsection{Stream-based Classification}
\label{sec:online_classification_result}

The stream-based \textsc{ml} models are implemented using the River library: \textsc{gnb}\footnote{Available at \url{https://riverml.xyz/dev/api/naive-bayes/GaussianNB}, May 2025.}, \textsc{lr}\footnote{Available at \url{https://riverml.xyz/0.11.1/api/linear-model/LogisticRegression}, May 2025.}, \textsc{alma}\footnote{Available at \url{https://riverml.xyz/0.11.1/api/linear-model/ALMAClassifier}, May 2025.}, \textsc{hatc}\footnote{Available at \url{https://riverml.xyz/0.13.0/api/tree/HoeffdingAdaptiveTreeClassifier}, May 2025.} and \textsc{arfc}\footnote{Available at \url{https://riverml.xyz/0.11.1/api/ensemble/AdaptiveRandomForestClassifier}, May 2025.}. Listings \ref{log_conf}, \ref{alma_conf}, \ref{hatc_conf} and \ref{arfc_conf} detail the ranges used to optimize the hyper-parameters with the selected values in bold for the \textsc{lr}, \textsc{alma}, \textsc{hatc} and \textsc{arfc} algorithms, respectively\footnote{Note that the latter hyper-parameter optimization process does not apply to the baseline \textsc{gnb} algorithm.}.

\begin{lstlisting}[frame=single,caption={\textsc{lr} hyperparameter configuration.},label={log_conf},emphstyle=\textbf,escapechar=ä]
l2 = [ä\bf{0.0}ä, 0.1, 1.0]
interceptlr = [ 0.001, ä\bf0.01ä, 0.1]
\end{lstlisting}
 
\begin{lstlisting}[frame=single,caption={\textsc{alma} hyperparameter configuration.},label={alma_conf},emphstyle=\textbf,escapechar=ä]
alpha = [ä\bf0.5ä, 0.7, 0.9]
B = [ä\bf{0.6}ä, 1.0, 1.4]
C = [1.0, ä\bf1.4ä, 1.8]
\end{lstlisting}

\begin{lstlisting}[frame=single,caption={\textsc{hatc} hyperparameter configuration.},label={hatc_conf},emphstyle=\textbf,escapechar=ä]
depth = [ä\bf{None}ä, 50, 200]
tiethreshold = [0.5, ä\bf0.05ä, 0.005]
maxsize = [ä\bf50ä, 100, 200]
\end{lstlisting}

\begin{lstlisting}[frame=single,caption={\textsc{arfc} hyperparameter configuration.},label={arfc_conf},emphstyle=\textbf,escapechar=ä]
models = [10, 50, ä\bf100ä]
features = [ä\bf{sqrt}ä, 5, 50]
lambda = [10, 50, ä\bf100ä]
\end{lstlisting}

Table \ref{tab:classification_results} compares the results of the three models. As it can be observed, \textsc{gnb} exhibits uniform values across the evaluation metrics within the \SI{70}{\percent} to \SI{80}{\percent} range. The \textsc{alma} classifier has improved regular results within the \SI{80}{\percent} to \SI{90}{\percent} range. More unstable results are those attained by \textsc{lr} (\textit{e.g.}, recall and \textsc{f}-measure of \textsc{ppd} absence). The latter also applies to \textsc{hatc}. Ultimately, \textsc{arfc} is the model that obtained the best results, surpassing the \SI{90}{\percent} threshold.

\begin{table*}[!htbp]
\centering
\footnotesize
\caption{\label{tab:classification_results}\textcolor{black}{Prediction results in streaming after the last sample (\#0 and \#1 stand for absence and presence of \textsc{ppd}, respectively).}}
\begin{tabular}{ccccccccccccc}
\toprule
 \bf Model & \bf Acc. & \bf AUC & \multicolumn{3}{c}{\bf Precision} & \multicolumn{3}{c}{\bf Recall} & \multicolumn{3}{c}{\bf F-measure} &  \bf Time (s)\\
\cmidrule(lr){4-6}
\cmidrule(lr){7-9}
\cmidrule(lr){10-12}
  & & & Macro & \#0 & \#1 & Macro & \#0 & \#1 & Macro & \#0 & \#1 & \\
\midrule

\textsc{gnb} & 80.05 & 77.26 & 78.32 & 73.35 & 83.28 & 77.26 & 67.88 & 86.65 & 77.72 & 70.51 & 84.93 & 0.26\\
\textsc{lr} & 78.52 & 70.94 & 81.88 & 87.18 & 76.58 & 70.94 & 45.51 & 96.38 & 72.57 & 59.80 & 85.35 & 0.15\\
\textsc{alma} & 86.85 & 85.70 & 85.51 & 80.91 & \bf 90.11 & 85.70 & \bf 81.84 & 89.56 & 85.60 & 81.37 & 89.83 & 0.10\\ 
\textsc{hatc} & 75.35 & 70.00 & 73.65 & 70.10 & 77.20 & 70.00 & 52.01 & 87.99 & 70.98 & 59.71 & 82.24 & 0.49\\
\textsc{arfc} & \bf 91.40 & \bf 88.82 & \bf 92.32 & \bf 94.58 & 90.06 & \bf 88.82 & 80.11 & \bf 97.52 & \bf 90.19 & \bf 86.75 & \bf 93.64 & 34.28\\

\bottomrule
\end{tabular}
\end{table*}

As shown in Table \ref{tab:dataset_distribution}, in the experimental data, a significant increase exists in the number of positive cases compared to the control cases. To analyze if the proposed solution is affected by data imbalance that may result in potential false positive cases, Table \ref{tab:classification_results_balanced} details the results obtained in a balanced setting. For this purpose, we applied balanced resampling using the down-sampling technique for the minority class (\textit{i.e.}, 523 cases with and without \textsc{ppd}). As expected, the results are better in the balanced scenario. In the case of the \textsc{gnb} and \textsc{hatc} models, only the precision in detecting the presence of \textsc{ppd} is affected, while for \textsc{lr}, it is the recall. In contrast, the \textsc{alma} model attains more stable values. Similarly, the \textsc{arfc} has slight variations in the evaluation metrics, proving its robustness compared to the other models. More importantly, the evaluation results are consistent across the unbalanced and balanced settings. Thus, our proposal avoids the potential false positive problem, thanks to the ensemble strategy in which the prediction is obtained through majority voting.

\begin{table*}[!htbp]
\centering
\footnotesize
\caption{\label{tab:classification_results_balanced}\textcolor{black}{Prediction results in the balanced setting in streaming after the last sample (\#0 and \#1 stand for absence and presence of \textsc{ppd}, respectively).}}
\begin{tabular}{ccccccccccccc}
\toprule
 \bf Model & \bf Acc. & \bf AUC & \multicolumn{3}{c}{\bf Precision} & \multicolumn{3}{c}{\bf Recall} & \multicolumn{3}{c}{\bf F-measure} &  \bf Time (s)\\
\cmidrule(lr){4-6}
\cmidrule(lr){7-9}
\cmidrule(lr){10-12}
  & & & Macro & \#0 & \#1 & Macro & \#0 & \#1 & Macro & \#0 & \#1 & \\
\midrule

\textsc{gnb} & 78.64 & 78.64 & 79.09 & 82.71 & 75.47 & 78.64 & 72.41 & 84.87 & 78.56 & 77.22 & 79.89 & 0.18\\
\textsc{lr} & 78.66 & 78.65 & 79.23 & 83.30 & 75.17 & 78.65 & 71.65 & 85.66 & 78.55 & 77.03 & 80.07 & 0.11\\
\textsc{alma} & 84.59 & 84.59 & 84.59 & 84.51 & 84.67 & 84.59 & 84.67 & 84.51 & 84.59 & 84.59 & 84.59 & 0.07\\
\textsc{hatc} & 75.10 & 75.10 & 77.29 & 85.03 & 69.55 & 75.10 & 60.92 & 89.27 & 74.59 & 70.98 & 78.19 & 0.34\\
\textsc{arfc} & \bf 90.13 & \bf 90.13 & \bf 90.18 & \bf 91.49 & \bf 88.87 & \bf 90.13 & 
\bf 88.51 & \bf 91.76 & \bf 90.13 & \bf 89.97 & \bf 90.29 & 21.79\\
 
\bottomrule
\end{tabular}
\end{table*}

Finally, Table \ref{tab:results_comparison} compares these results to the most related competing work in the literature. Note that the work by \citet{sezgin2023clinical} is a non-\textsc{ml} study, and no classification metrics are provided. Taking into account the already mentioned theoretical differences (\textit{e.g.}, none performed online processing or provided explainability in natural language), our proposal is the first to address \textsc{ppd} detection combining \textsc{llm}s and \textsc{ml} and employing free-speech input data. 

The highest difference in accuracy is with the work by \citet{Andersson2021}, +\SI{17.90}{\percent} points. The performance improvement is greater regarding the \textsc{auc} metric (+\SI{21.68}{\percent} points compared to \citet{Duvvuri2022}), recall (+\SI{26.02}{\percent} points compared to \citet{Duvvuri2022}), and precision (+\SI{28.77}{\percent} points compared to \citet{Andersson2021}). The greater improvement is obtained in \textsc{f}-measure (+\SI{40.19}{\percent} points) compared to \citet{Liu2023}. On the contrary, the slightest difference is with the research by \citet{Amit2021} and \citet{zhang2025interpretable} for the \textsc{auc} metric, even though for the last work, the difference in accuracy exceeds \SI{10}{\percent} points in favour of our proposal. With regard to the study by \citet{qi2025prediction}, the improvements are +\SI{14.52}{\percent} points, +\SI{9.07}{\percent} points, and +\SI{18.69}{\percent} points for precision, recall and \textsc{f}-measure, respectively.

\begin{table}[!htbp]
\centering
\caption{\label{tab:results_comparison}\textcolor{black}{Comparison with the results of relevant research.}}
\begin{tabular}{lccccc}
\toprule \textbf{Authorship} & \textbf{Accuracy} & \textbf{AUC} & \textbf{Precision} & \textbf{Recall} & \textbf{F-measure}\\ \midrule

\citet{Amit2021} & - & 84.40 & - & 78.00 & -\\

\citet{Andersson2021} & 73.50 & 80.95 & 63.55 & 73.50 & -\\

\cite{Duvvuri2022} & 88.21 & 67.14 & - & 62.80 & -\\

\cite{Liu2023} & 74.20 & - & - & 73.10 & 50.00 \\

\citet{qi2025prediction} & - & 85.80 & 77.80 & 79.75 & 71.50\\

\citet{zhang2025interpretable} & 81.30 & 84.90 & 78.65 & 79.00 & - \\\midrule

Proposed & \bf 91.40 & \bf 88.82 & \bf 92.32 & \bf 88.82 & \bf 90.19\\
\bottomrule
\end{tabular}
\end{table}

\subsection{Stream-based Explainability}
\label{sec:explainability_results}

To generate the counterfactual explanations, we use samples whose prediction using the \texttt{\small Predict\_Proba\_One} function exceeds \SI{80}{\percent}. These samples are modified by introducing variations in each feature until the prediction changes to above \SI{50}{\percent}. Additionally, these permutations are carefully designed to avoid changes that contradict the model's logic.

Listing \ref{counter_factual} shows the result of two samples evaluated using the counterfactual explanations algorithm. The first example shows the variations required for the transition of \textsc{ppd} positive prediction to absence. Conversely, the second example illustrates the opposite case. The most important features are those that modify the classifier prediction above the \SI{50}{\percent} threshold (in bold), while the remaining features exhibit negligible influence on the output.

\begin{lstlisting}[frame=single,caption={Counterfactual explanations examples.},label={counter_factual},emphstyle=\textbf,escapechar=ä]

Presence of PPD (84.26 %)  to Absence of PPD (52.03 %)

ä\bf Age ä:  30-35 -> 35-40
ä\bf Baby bonding issues ä: Sometimes -> No
Concentration and decision-making problems:  No 
Feeling sad or tearful: Sometimes 
Feeling guilty: No 
ä\bf Irritable towards the baby or the partner ä: Yes -> No
ä\bf Overeating or loss of appetite ä:  Yes -> Sometimes
Suicide behavior: Unwilling to disclose
ä\bf Trouble sleeping ä: Sometimes -> No

Absence of PPD (83.75 %)  to Presence of PPD (57.82 %)

Age: 30-35
ä\bf Baby bonding issues ä: No -> Yes
ä\bf Concentration and decision-making problems ä:  No -> Yes
Feeling sad or tearful: Yes 
ä\bf Feeling guilty ä: Yes -> Sometimes 
Irritable towards the baby or the partner: Sometimes
Overeating or loss of appetite:  Yes
ä\bf Suicide behavior ä: No -> Yes
Trouble sleeping: Often

\end{lstlisting}

\begin{figure*}[!htbp]
 \centering
 \includegraphics[scale=0.22]{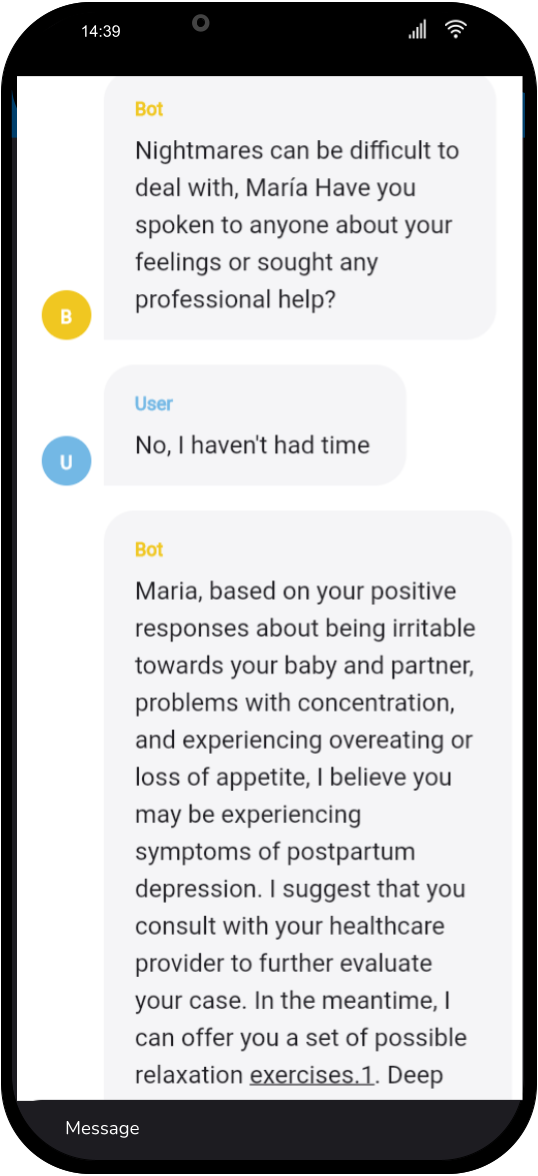}
 \caption{Explainability information inserted in a real conversation.}
 \label{fig:dasboard}
\end{figure*}

Figure \ref{fig:dasboard} depicts a real conversation with an end user, including case-specific interpretability data. Note that the system provides information about the most relevant features in the decision process. The explainability outcomes are inserted as part of the conversation, and finally, recommendations are generated employing the third prompt in Table \ref{tab:prompts} to reduce the stress level when \textsc{ppd} is detected.

\section{Conclusion}
\label{sec:conclusion}

Women's mental and physical well-being is being significantly affected by pregnancy and childbirth. However, the current solutions compromise the early detection of \textsc{ppd}. Surprisingly, despite the severity of \textsc{ppd} in our society, the number of current studies on this topic is scarce when it comes to the newest \textsc{ai} techniques (\textit{e.g.}, \textsc{llm}s, \textsc{xai}, online processing).

Consequently, our work contributes to an intelligent \textsc{ppd} screening system from spontaneous speech that combines \textsc{nlp}, \textsc{ml}, and \textsc{llm}s towards an affordable, real-time, and non-invasive procedure. Moreover, it addresses the black box problem since the predictions are explained with feature importance and natural language. Results obtained are \SI{90}{\percent} on \textsc{ppd} detection for all evaluation metrics, outperforming the competing solutions in the literature. Ultimately, our solution has the potential to help healthcare practitioners detect at-risk patients by assisting them during decision-making through precise estimates.

The healthcare applicability of our work is directed to maternal healthcare settings, specifically in prenatal clinics and community-based maternal programs. The system could be integrated into mobile health platforms, allowing healthcare professionals to use it in real-time during patient visits. The system’s output would serve as a decision-support tool, aiding clinicians in screening for early adverse signs, prioritizing patients for triage, or even guiding intervention when risk levels surpass predefined thresholds. Such integration enhances early detection and helps optimize clinical resources by focusing on high-risk cases, particularly in underserved or resource-limited settings. From the patient’s perspective, the proposed system offers several key advantages. First, enabling earlier and more accurate detection of potential health risks through continuous monitoring can lead to timely clinical intervention. Second, integrating the system into health platforms may increase engagement and adherence to medical recommendations. Third, in low-resource or rural areas with limited access to specialized maternal care, the system can be a remote monitoring tool, reducing unnecessary clinic visits while ensuring that high-risk cases receive prompt attention.

In practical use cases, informed consent must be explicitly obtained from all users before collecting or processing data. This involves communicating the purpose of the system, the nature of the data being collected, how it will be used, stored, and protected, and providing users with the option to withdraw at any time. Regarding privacy, on-device processing and strong encryption protocols for any transmitted data ensure no personally identifiable information is compromised. To ensure accountability, system decisions, particularly those that may influence clinical care, are always intended to be reviewed by qualified health professionals. In this regard, explainability modules allow users and clinicians to understand the rationale behind outputs. Finally, evaluation of model performance across diverse groups of users and continual updates with new data will help to mitigate bias.

While the system strongly emphasizes explainability, we recognize the critical importance of ensuring a positive and sensitive user experience, particularly for postpartum women experiencing emotional distress. In this line, our proposal is designed to operate as a supportive decision aid rather than a diagnostic tool. The system does not store identifiable personal data, minimizing privacy risks. In future work, we plan to analyze the empathetic capabilities of the solution with real experimental data gathered from end users. Moreover, we plan to consider other languages like Spanish and how new features influence \textsc{ppd} prediction. Furthermore, alternative classification perspectives (\textit{e.g.}, reinforcement learning, unsupervised classification) will be considered. The computing and storage load associated with the current version of the system will be analyzed to ensure its effective performance. Consideration should also be given to measure the system's acceptability and safety. Ultimately, clinical experts will be invited to validate the interpretation of features or chatbot prompts to address the limitations in representativeness or generalizability of the datasets.

\section*{Disclosure Statement}

The authors have no competing interests to declare relevant to this article's content.

\section*{Funding}

This work was partially supported by Xunta de Galicia grants ED481B-2022-093 and ED481D 2024/014.

\section*{ORCID}

Silvia García Méndez \url{https://orcid.org/0000-0003-0533-1303}, Francisco de Arriba Pérez \url{https://orcid.org/0000-0002-1140-679X}.

\section*{Authorship Contribution Statement}

All authors have read and approved the final version of the manuscript. \textbf{Silvia García-Méndez}: Conceptualization, Methodology, Software, Validation, Formal analysis, Investigation, Resources, Data Curation, Writing - Original Draft, Writing - Review \& Editing, Visualization, Funding acquisition. \textbf{Francisco de Arriba-Pérez}: Conceptualization, Methodology, Software, Validation, Formal analysis, Investigation, Resources, Data Curation, Writing - Original Draft, Writing - Review \& Editing, Visualization, Funding acquisition.

\section*{Data and Code Availability Statement}

Experimental data is publicly available at \url{https://doi.org/10.5281/zenodo.14049633} and
\url{https://www.kaggle.com/datasets/parvezalmuqtadir2348/postpartum-depression}. The code is publicly available at \url{https://github.com/nlpgti/postpartum_stress}.

\bibliography{2_bibliography}

\end{document}